\pdfoutput=1

\documentclass[11pt]{article}

\usepackage{EMNLP2023}

\usepackage{times}
\usepackage{latexsym}

\usepackage[T1]{fontenc}

\usepackage[utf8]{inputenc}

\usepackage{microtype}

\usepackage{inconsolata}
\usepackage{graphicx}
\usepackage{hyperref}

\title{Translation Canvas: An Explainable Interface to Pinpoint and Analyze Translation Systems}

\author{
    Chinmay Dandekar\textsuperscript{\ddag}, Wenda Xu\textsuperscript{\ddag}, Xi Xu\textsuperscript{\dag}, Siqi Ouyang\textsuperscript{\dag}, Lei Li\textsuperscript{\dag} \\
    \textsuperscript{\ddag}University of California, Santa Barbara
    \textsuperscript{\dag}Carnegie Mellon University\\
    \texttt{\{cdandekar, wendaxu\}@ucsb.edu} \\
    \texttt{\{siqiouya, xixu, leili\}@cs.cmu.edu} 
}

\begin{document}
\maketitle

\begin{abstract}With the rapid advancement of machine translation research, evaluation toolkits have become essential for benchmarking system progress. Tools like COMET and SacreBLEU offer single quality score assessments that are effective for pairwise system comparisons. However, these tools provide limited insights for fine-grained system-level comparisons and the analysis of instance-level defects. To address these limitations, we introduce \textbf{Translation Canvas}, an explainable interface designed to pinpoint and analyze translation systems' performance: 1) Translation Canvas assists machine translation researchers in comprehending system-level model performance by identifying common errors (their frequency and severity) and analyzing relationships between different systems based on various evaluation metrics. 2) It supports fine-grained analysis by highlighting error spans with explanations and selectively displaying systems' predictions. According to human evaluation, Translation Canvas  demonstrates superior performance over COMET and SacreBLEU packages under enjoyability and understandability criteria. 



    
    
\end{abstract}

\section{Introduction}

As natural language processing (NLP) technologies evolve, the need for precise and detailed analysis of model outputs has become increasingly critical. Despite significant advancements in translation models, a gap remains in the tools available for researchers to thoroughly evaluate and interpret these models' predictions. This issue is particularly acute in the context of translation research, where understanding the nuances of model errors and performance is vital for further improvements.

Translation model developers excel in creating sophisticated algorithms, but often face challenges when it comes to conducting fine-grained analysis of model predictions. Moreover, tools designed to facilitate such analysis typically lack the flexibility and specificity required for detailed evaluation at the instance level. This disconnect underscores the necessity for an integrated solution that combines comprehensive model evaluation with user-friendly interfaces and advanced analytical capabilities.

Existing approaches to model evaluation often focus on high-level metrics such as BLEU or COMET scores, which, while useful, do not provide the granularity needed to identify specific areas of improvement like stylistic errors and incorrect word choices. Moreover, the process of manually analyzing individual model predictions is time-consuming and prone to error. Additionally, analyzing predictions in a language direction that translation researchers are unfamiliar creates a language barrier and hinders model improvement. As translation models continue to grow in complexity, the demand for a more sophisticated, streamlined approach to model evaluation and error analysis has never been higher.

Translation Canvas addresses these challenges by offering a comprehensive toolkit designed specifically for translation researchers. It provides a dashboard that displays the distribution of common errors, instance-level performance and system-level performance for each model. This helps the user identify specific areas of improvement in their models, and identify gaps in model performances. The system also provides fine-grained analysis by displaying instances. The system visually highlights erroneous text spans and provides natural language explanation explaining the error. Users can also construct complex search queries to filter instances and conduct targeted analysis. Our evaluation shows that users find the system to be useful, enjoyable and as easy to use as command-line evaluation tools. 

Our contributions are summarized as follows:

\begin{itemize}
    \item Our tool demonstrates system-level performance by identifying error and score distribution and analyzing relationships between different systems
    \item We display instance-level performance by highlighting error spans with explanations and selectively  displaying systems’ predictions
    \item Our human evaluation shows that Translation Canvas's superior usability and enjoy-ability compared to prior systems like SacreBLEU and COMET
\end{itemize}

In this paper, we present Translation Canvas and demonstrate how it meets the critical needs of translation researchers. We outline its key features, including instance-level analysis, error classification, and advanced search capabilities, and illustrate how these tools can be leveraged to gain deeper insights into model behavior. Through detailed examples and use cases, we show how Translation Canvas transforms the process of translation model evaluation, making it more efficient, accurate, and insightful. 

\section{Related Works}

Recent years have seen a growing interest in developing tools and frameworks for comprehensive evaluation and analysis of NLP models, particularly in the domain of machine translation. These efforts aim to provide researchers with deeper insights into model performance, error patterns, and areas for improvement.

Many evaluation metrics have been proposed to measure, evaluate, and explain machine translation. These metrics include both automatic and human evaluation methods. Automatic metrics, such as BLEU \cite{papineni-etal-2002-bleu}, METEOR \cite{denkowski2014meteor}, and ROUGE \cite{lin2004rouge}, measure the quality of translations by comparing them to human references, focusing on aspects like n-gram overlap and recall. More advanced metrics like BERTScore \cite{zhang2019bertscore}, COMET \cite{rei-etal-2020-comet} and SEScore \cite{xu-etal-2022-errors, xu-etal-2023-sescore2} use learned techniques to assess translation quality by comparing sentence embeddings. InstructScore \cite{xu2023instructscoreexplainabletextgeneration} leverages large language models to provide error classifications and explanations on an instance-level. Human evaluation methods, such as the Multidimensional Quality Metrics (MQM) \cite{lommel2014multidimensional} framework, involve human judges assessing translation quality based on criteria like fluency and adequacy. These methods provide nuanced feedback but require effort from the model developer to interpret and process in order to make effective diagnosis of models. In addition, each metric must be evaluated and understood separately, making it harder to leverage the multiple metrics to identify core issues with models.

Some visual frameworks have been proposed to provide model developers with a unified interface to evaluate and debug models. ExplainaBoard \cite{liu2021explainaboardexplainableleaderboardnlp} 
offers an explainable leaderboard for NLP tasks that provides fine-grained analyses of model performance. While ExplainaBoard offers valuable insights, Translation Canvas builds upon this concept by providing more specialized tools for translation model analysis, an instance-level analysis of errors. MT-Telescope \cite{rei-etal-2021-mtelescope} provides an evaluation and visualization platform for machine translation systems that supports comparison of models, dynamically filtering content and visualizations that enhance model comparisons. Although MT-Telescope provides a fantastic platform for model comparison, Translation Canvas provides a more flexible content filtering system, allowing users to join requests to produce a complex search query. While MT-Telescope is focused on model comparisons, Translation Canvas is flexible in the number of models it can compare, as well as having the option to analyze a model by itself.
MATEO \cite{vanroy-etal-2023-mateo} provides a suite of evaluation metrics for machine translation, and visualization for model performance via a user-friendly web application interface. While MATEO lets users easily evaluate their models on a wide variety of metrics, Translation Canvas provides users the ability to do fine-grained analysis with natural language error explanations, as well as an advanced search system.

Translation Canvas builds upon these existing works by integrating comprehensive evaluation metrics, fine-grained error analysis, and an intuitive user interface specifically designed for translation researchers. Our system uniquely combines features like advanced search functionality, model comparison dashboards, and instance-level analysis, addressing the need for a specialized toolkit in the field of machine translation evaluation.

\begin{figure*}[t]
    \centering
    \includegraphics[width=1\linewidth]{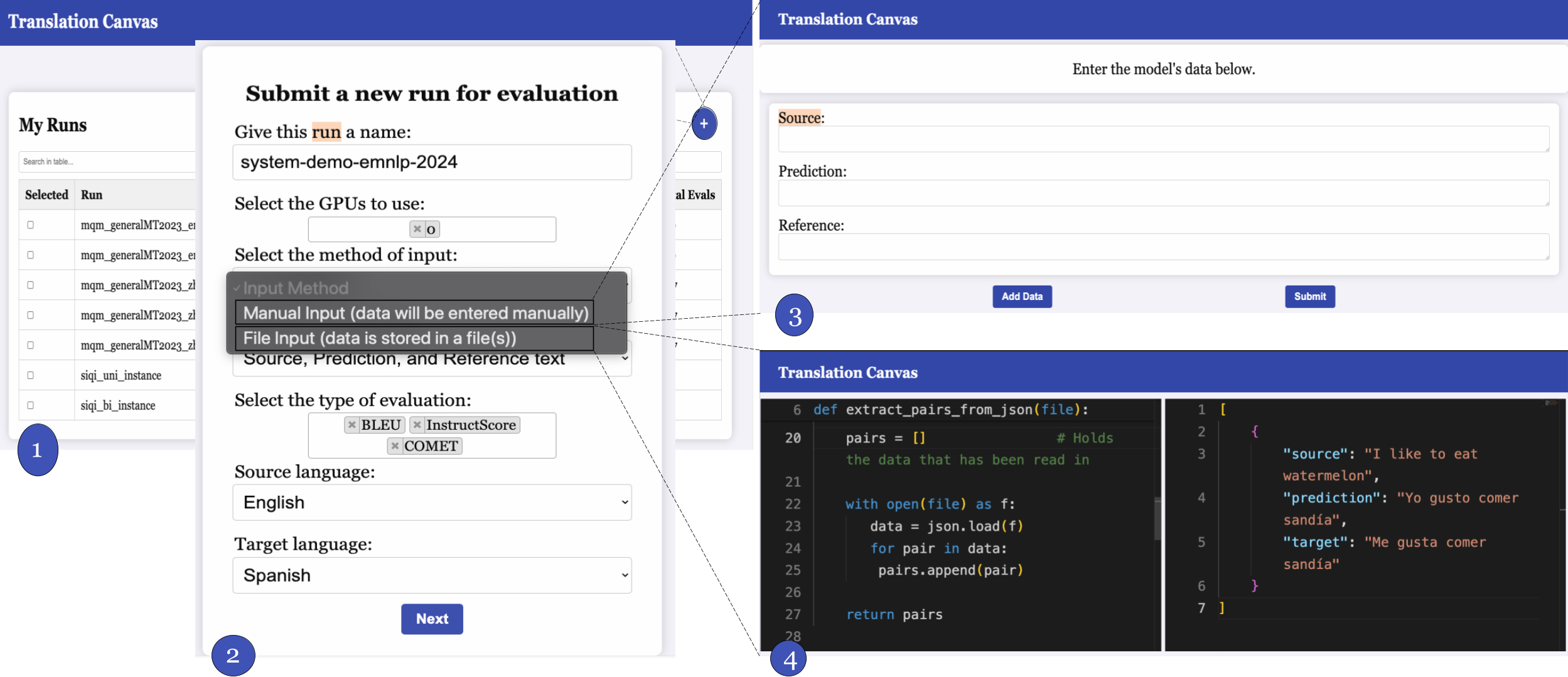}
    \caption{This is the workflow for submitting a model's instance for evaluation in Translation Canvas. The user can choose from (2) to manually input the instances (3) or extract the instances from a file (4).}
    \label{fig:submit-workflow}
\end{figure*}

\section{Translation Canvas}

Translation Canvas is implemented using Python Flask. It uses a Flask backend and Jinja templates for frontend with a DuckDB connection. It is distributed over pip\footnote{\href{https://pypi.org/project/translation-canvas/}{https://pypi.org/project/translation-canvas/}} and is available for use to everyone under the MIT open-source license.

\begin{figure*}[t]
    \centering
    \includegraphics[width=1\linewidth]{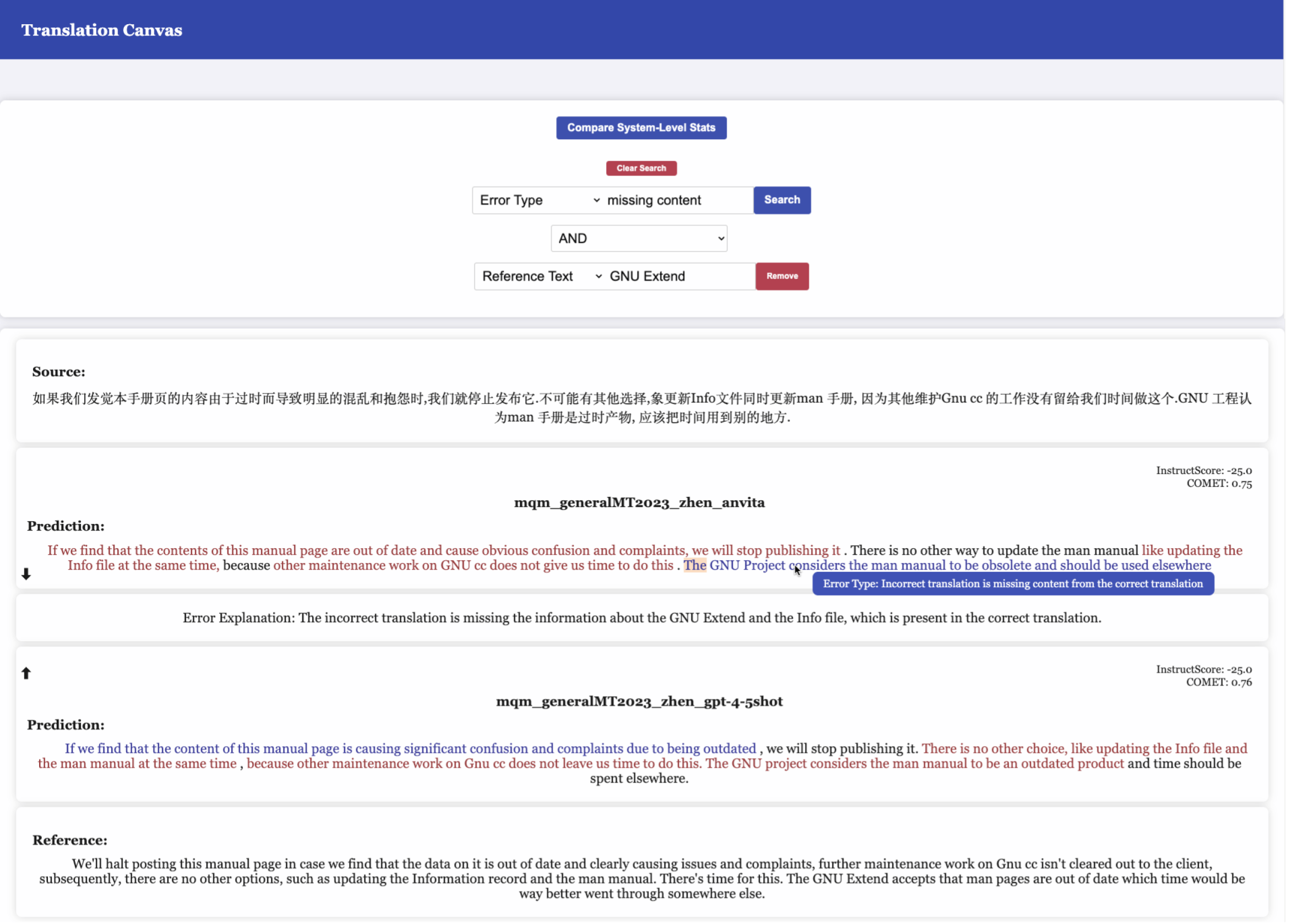}
    \caption{Instance-level comparison of GPT4-5shot and ANVITA  model evaluation}
    \label{fig:instance-use-case}
\end{figure*}

\begin{figure*}[t]
    \centering
    \includegraphics[width=1\linewidth]{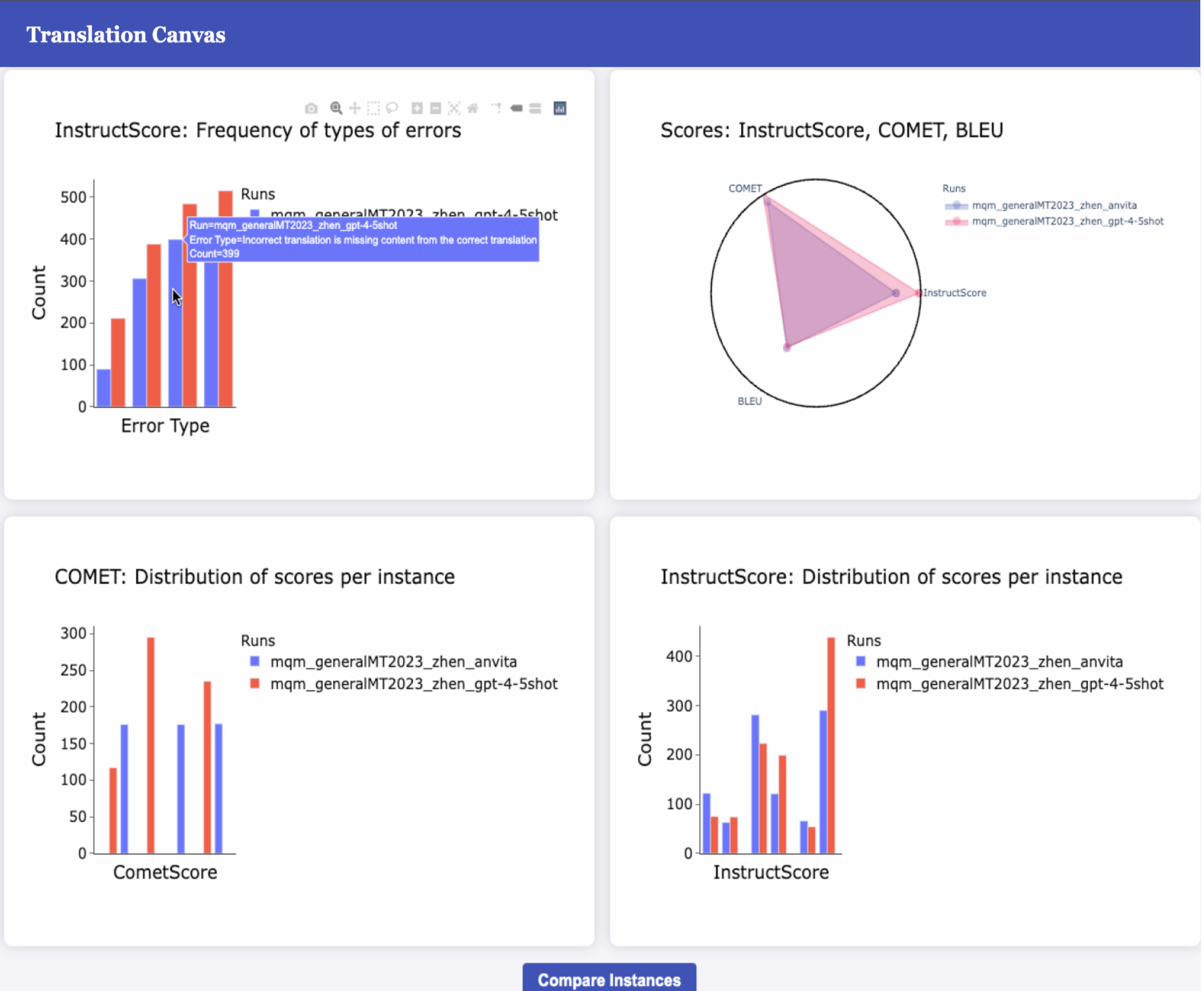}
    \caption{System-level comparison of GPT4-5Shot and ANVITA model evaluation}
    \label{fig:system-use-case}
\end{figure*}


\subsection{Evaluating instances}

Translation Canvas features a customizable, flexible and easy method for submitting instances for evaluation. It allows the user to specify the evaluations that they want the system to run, the GPUs they want to run the evaluations on, and if they want to submit references and sources in the instances. 

Translation Canvas currently supports 3 evaluation metrics: 
\begin{itemize}
    \item InstructScore \cite{xu2023instructscoreexplainabletextgeneration}, an explainable evaluation metric for text generation that uses a fine-tuned LLaMA model to produce both a score and a detailed diagnostic report per error.
    \item BLEU \cite{papineni-etal-2002-bleu}, a metric used to evaluate the quality of machine-generated text by measuring the overlap of n-grams between the generated text and one or more reference translations.
    \item COMET \cite{rei-etal-2020-comet}, a metric that uses neural models to evaluate machine translation quality by predicting human judgment scores.
\end{itemize}

To accommodate for all types of instance input, we allow the user to input instances using 2 methods: \\
\\
\textbf{Manual Input}\ \ The system allows the user to manually input source, prediction, and reference text. This option is intended for quick evaluations of a couple of predictions. Figure \ref{fig:submit-workflow} shows an example of the manual input page. \\
\\
\textbf{File Input}\ \ The system accepts text based files of all formats. This allows the user to submit instances with just a small amount of post-processing to submit to the system. To be able to accommodate any text-based file and extract the source, prediction and reference text, the system provides the user with an integrated development environment. The system asks the user to write a small function that reads the text file appropriately and extracts the relevant information from the file. By doing this, the system can accept any text-based file. Figure \ref{fig:submit-workflow} shows an example of the file input page.

\subsection{Instance Analysis}

Translation Canvas' central feature is its integrated instance analysis page. The page renders the source, prediction, and reference text together. If more than one model's instances are being rendered, the instances are grouped by reference and source for easy comparison between the models' predictions. \\
\\
\textbf{InstructScore}\ \  If the models are evaluated with InstructScore, then the predictions are rendered with error information. For each prediction-reference pair, InstructScore provides error information about the prediction including error type, scale, location, and explanation. 

The erroneous section of the prediction text is highlighted with a red or orange color. This allows users to easily identify which subsections of the prediction are causing the issues. Red text signifies a major error, while orange text signifies a minor error. This helps the user easily identify the distribution of errors in an instance. In addition, the instance level COMET and InstructScore are displayed, to give the user an understanding of how accurate each instance is to the reference.

When the user hovers over the red or orange text, a tooltip appears. This contains helpful information about the error types, scale, and explanation made in the prediction. This is especially helpful when the user is not familiar with a language direction that they are evaluating. In Figure \ref{fig:instance-use-case}, we can see the mouse hovering over the segment 'GNU Project considers$\dots$' in blue, which displays the reason that segment is wrong in the translation. \\
\\
\textbf{Comparison}\ \ Translation canvas supports comparing the instances of models. The system groups instances by source and reference text. With this, the user is able to easily compare models to identify the difference between predictions, including identifying places where one model made an error while the other didn't. This helps the user understand why a model is doing worse than a reference model. The order the predictions are displayed is sorted by the quality of the prediction. The predictions also have up and down arrows on them, where users can re-rank the order of predictions based on quality. Given user permission, we collect this user feedback, including source, reference, model output, and the user ranking for further improvements to the system. The user can choose to revoke permission at any time. In Figure \ref{fig:instance-use-case}, we show fine-grained analysis of the machine translation models GPT4-5Shot \cite{gpt45shot} and ANVITA \cite{kocmi-EtAl:2023:WMT}, submitted to the WMT 2023 General Translation Task \cite{kocmi-EtAl:2023:WMT}, being compared for the Chinese to English language direction.

\subsection{Search}

Translation Canvas supports a powerful search feature at the instance level that allows users to construct a complex query. Users can search by the following categories:
\begin{itemize}
    \item Errors (Type, Scale, and Explanation)
    \item Text (Source, Prediction, and Reference)
    \item Languages (Source and Target)
\end{itemize}
The search text can be used to filter for a specific instance based on the categories listed above. The search text also supports SQL style regular expressions, allowing users flexibility when searching. They can also join together an arbitrary number of queries together with a choice of 'AND', 'OR', and 'AND NOT' conjunctions. This gives the user a lot of flexibility when searching for instances to analyze a model's output.

When searching by error type, location, scale or explanation, Translation Canvas will highlight the selected errors in blue. This helps the user easily identify the errors that are being searched for. Figure \ref{fig:instance-use-case} shows an example of searching for the error type 'missing content' and reference text that contains 'GNU Extend'.

\subsection{Dashboard}

Translation Canvas features a dashboard that allows an analysis of models on the corpus level. It presents histograms on the distribution of InstructScore, the distribution of COMET, the distribution of the error types of a model, and the corpus-level BLEU, COMET and InstructScore of the models. This information allows users to identify the model's biggest weaknesses, so that the user can investigate them on a closer level.  \\
\\
\textbf{Comparison} \ \ The dashboard allows for comparisons of models. It displays each model's statistics side-by-side, which allows the user to identify the major gaps in a model's performance compared to other models. It also allows the user to understand the consistency of each model, based on the distribution of instance level COMET and InstructScore scores. This allows the user greater insight into each model. Figure \ref{fig:system-use-case} shows an example of such a dashboard.

\section{Use Case}

Here, we briefly showcase a use case for Translation Canvas to evaluate models and identify gaps between performance. For this use case, we use the ANVITA \cite{kocmi-EtAl:2023:WMT} and GPT4-5Shot \cite{gpt45shot} machine translation models, submitted to the WMT 2023 General Translation Task \cite{kocmi-EtAl:2023:WMT}. After extracting and evaluating the instances, we can start by comparing the ANVITA and GPT4-5shot instances at a system level.

It is clear from the dashboard in Figure \ref{fig:system-use-case}  that GPT4-5Shot has a far more favorable distribution of instance level InstructScore (bottom right) and also makes far fewer errors of the type "Incorrect translation is missing content from the correct translation". This implies that ANVITA is leaving out content from the translation at a much higher frequency than GPT4-5Shot is. We can investigate this further by switching to an instance-level view.

To filter the instances that we are interested in analyzing, we can use the search bar to search by error type and find only instances with error type "missing content", which will find all instances with those errors. In Figure \ref{fig:instance-use-case}, we see an instance with predictions from the 2 models. The text written in blue contains errors that we are looking for. When the hovering the mouse over the blue span of the text, we see the natural language explanation of the error.

Note that the error span for ANVITA is at the very end, and the explanation shows that ANVITA is missing content at the very end of the translation. This could help the model developer understand the weaknesses of ANVITA, such as possibly a tendency to truncate or omit information at the end of longer sentences. This insight could guide improvements to the model's handling of sentence endings or its ability to maintain context throughout longer translations.

\section{Evaluation}

To assess the effectiveness of Translation Canvas, we conducted a user evaluation study with participants who have substantial expertise in machine translation and knowledge of existing MT metrics. Our evaluation focused on two key aspects: instance-level analysis and system-level analysis.

For instance-level analysis, Translation Canvas received high marks, with expert users rating it 4/5 for both enjoyability and usability. Participants appreciated the highlight of error types and the quick analysis process. A particularly valuable feature was the elimination of the need for forward translation to understand unfamiliar languages, which even experienced users found beneficial. 

The system-level analysis features were also well-received by our expert evaluators, with enjoyability rated at 4/5 and usability at a perfect 5/5. Participants found the graph presentations appealing and particularly valued the sorted error types, which saved time in fine-grained analysis. The support for multi-system analysis was highlighted as a key usability feature. 

We also benchmarked the time required for non-experienced users, who have no prior knowledge of existing machine translation evaluation systems, to learn and use Translation Canvas compared to existing tools. Our findings showed that these first-time users took an average of 10 minutes to learn and use Translation Canvas on a custom dataset. This was comparable to the time needed for SacreBLEU \cite{post-2018-call} (10 minutes) and faster than COMET (15 minutes).

These results demonstrate that Translation Canvas provides an intuitive interface accessible to users without prior MT evaluation experience. The system's ability to match or exceed the learnability of established tools, while offering more comprehensive analysis features, indicates an effective balance between advanced functionality and user-friendly design. This combination of accessibility and depth potentially addresses a significant need in the MT research community for tools that facilitate both rapid onboarding and sophisticated analysis.

\section{Limitations and Future Work}

While these metrics provide valuable quantitative insights, they may not fully capture the nuanced aspects of translation quality that human expert evaluations could offer. The potential discrepancy between automatic and human evaluations underscores the need for a more comprehensive assessment approach.
To address this limitation, we have implemented a re-ranking feature in the fine-grained analysis interface. This feature allows users proficient in both source and target languages to manually adjust the ranking of predictions according to their expert judgment. This user-driven re-ranking serves a dual purpose: it provides immediate value to users seeking more accurate rankings and generates valuable data for future improvements. 

The collection of this user-generated re-ranking data presents an opportunity for future work. We intend to leverage this dataset to refine our ranking algorithms and enhance our evaluation methodologies. By incorporating human expertise into our automated systems, we aim to bridge the gap between automatic metrics and human judgment, potentially leading to more robust and reliable evaluation techniques in machine translation research.
\bibliography{anthology,custom}
\bibliographystyle{acl_natbib}

\end{document}